%% file: main.tex
\documentclass[10pt,twocolumn,letterpaper]{article}

\usepackage[pagenumbers]{cvpr} % To force page numbers, e.g. for an arXiv version

\input{preamble}

\definecolor{cvprblue}{rgb}{0.21,0.49,0.74}
\usepackage[pagebackref,breaklinks,colorlinks,citecolor=cvprblue]{hyperref}

\def\paperID{*****} % *** Enter the Paper ID here
\def\confName{CVPR}
\def\confYear{2024}

\title{Map-Assisted Remote-Sensing Image Compression at Extremely Low Bitrates}

\author{Yixuan Ye, Ce Wang, Wanjie Sun\thanks{Corresponding author}, Zhenzhong Chen\\
School of Remote Sensing and Information Engineering, Wuhan University\\ Wuhan 430079, China\\
\tt\small {\{yeyixuan, cewang, sunwanjie, zzchen\}@whu.edu.cn}
}
\begin{document}
\maketitle

\begin{abstract}
Remote-sensing (RS) image compression at extremely low bitrates has always been a challenging task in practical scenarios like edge device storage and narrow bandwidth transmission. Generative models including VAEs and GANs have been explored to compress RS images into extremely low-bitrate streams. However, these generative models struggle to reconstruct visually plausible images due to the highly ill-posed nature of extremely low-bitrate image compression. To this end, we propose an image compression framework that utilizes a pre-trained diffusion model with powerful natural image priors to achieve high-realism reconstructions. However, diffusion models tend to hallucinate small structures and textures due to the significant information loss at limited bitrates. Thus, we introduce vector maps as semantic and structural guidance and propose a novel image compression approach named Map-Assisted Generative Compression (MAGC). MAGC employs a two-stage pipeline to compress and decompress RS images at extremely low bitrates. The first stage maps an image into a latent representation, which is then further compressed in a VAE architecture to save bitrates and serves as implicit guidance in the subsequent diffusion process. The second stage conducts a conditional diffusion model to generate a visually pleasing and semantically accurate result using implicit guidance and explicit semantic guidance. Quantitative and qualitative comparisons show that our method outperforms standard codecs and other learning-based methods in terms of perceptual quality and semantic accuracy. The dataset and code will be publicly available at https://github.com/WHUyyx/MAGC.
\end{abstract}

\section{Introduction}
Satellites capture large volumes of high-resolution images of the Earth’s surface, resulting in an exponential growth of newly captured remote-sensing (RS) images every day. There exists a demand for extremely low-bitrate RS image compression techniques in practical scenarios, such as efficient long-term storage and analysis in cloud-based systems with storage limitations, and faster data transfer in emergency communication networks that may be constrained. The main challenge is to develop effective compression techniques that substantially reduce the bitrate while preserving essential RS image features, enabling accurate and reliable use of compressed images in various remote sensing applications \citep{tgrs2024}.
% High-throughput onboard hyperspectral image compression with ground-based CNN reconstruction
% Remote Sensing Image Compression Based on High-Frequency and Low-Frequency Components

In the past decades, the majority of image compression methods were designed upon frequency analysis, such as JPEG \citep{jpeg}, JPEG2000 \citep{jpeg2000}, BPG \citep{bpg} and VTM \citep{vtm}. These approaches generally employ a block-based transform coding framework and rely on hand-crafted features and heuristics to reduce redundancy in original images. At extremely low bitrates, traditional compression techniques can lead to significant loss of information. This loss can compromise the utility of the RS images for interpretation and analysis, as crucial features such as terrain, vegetation, and urban structures might be blurred or completely lost.
% The jpeg still picture compression standard
% Jpeg2000: Image compression fundamentals, standards and practice
% Bpg image format
% Versatile video coding
% Dynamic Kernel-Based Adaptive Spatial Aggregation for Learned Image Compression

\begin{figure*}[htb]
\centering
\includegraphics[width=\textwidth]{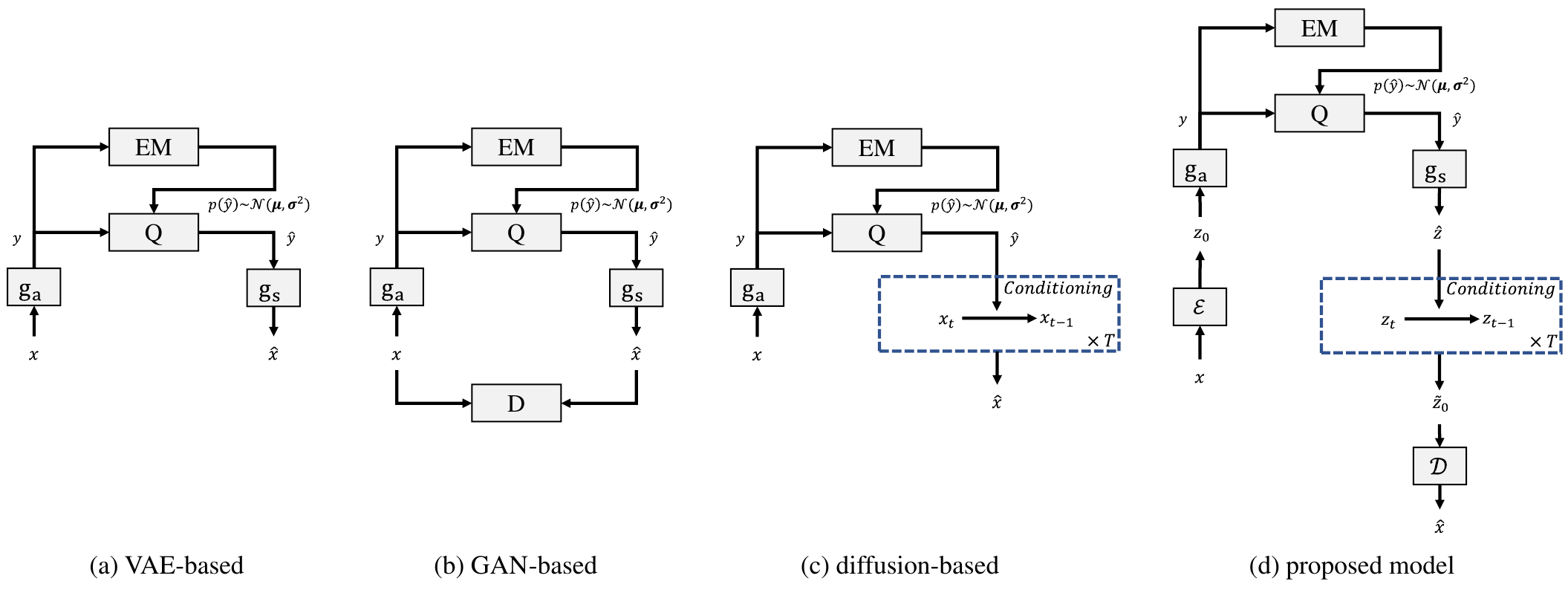}
\captionsetup{singlelinecheck=false} % 左对齐
\caption{Operational diagrams of learned image comprssion frameworks. Q denotes quantization. EM represents the entropy model. D denotes the discriminator. \begin{math}\mathcal{E}\end{math} and $\mathcal{D}$ denote the pre-trained SD VAE encoder and decoder used to transform data between pixel space and latent space.}
\label{fig1} 
\end{figure*} 

Recently, learned image compression methods have shown remarkable performance compared to traditional codecs \citep{channelwise, checkerboard, stf, elic, mixed}. Ballé et al. \citep{hyperprior} first proposed a learned image compression framework based on a variational autoencoder (VAE) architecture, as shown in Fig.\:\hyperref[fig1]{1(a)}. The encoder performs downsampling on the input image, yielding a compact latent representation that contains the most information about the original image. After quantization, the latent is transformed to a bitstream via entropy coding. On the decoding side, the image is reconstructed through the decoder. The entropy model is designed for distribution parameter estimation and probabilistic prediction for entropy coding to further reduce the bitrates. The learned image compression framework employs a data-driven, end-to-end training strategy, adaptively learning weights based on image contents. Consequently, the end-to-end learning approach outperforms traditional coding schemes where performance is limited by the handcrafted and fixed coding frameworks. However, these MSE-optimized VAE-based models tend to produce blurred results at low bitrates, limiting their practicality in such extreme low-bitrate scenarios.
% CHANNEL-WISE AUTOREGRESSIVE ENTROPY MODELS FOR LEARNED IMAGE COMPRESSION
% Checkerboard Context Model for Efficient Learned Image Compression
% The devil is in the details Window-based attention for image compression
% Efficient learned image compression with unevenly grouped space-channel contextual adaptive coding
% Learned Image Compression with Mixed Transformer-CNN Architectures
% VARIATIONAL IMAGE COMPRESSION WITH A SCALE HYPERPRIOR
% Neural datadependent transform for learned image compression

Considering the requirement of improving the perceptual quality of reconstructed images, researchers have proposed generative image compression frameworks based on generative adversarial network (GAN) \citep{gan} and diffusion model \citep{ddpm}, as shown in Fig.\:\hyperref[fig1]{1(b)} and Fig.\:\hyperref[fig1]{1(c)}. GAN-based methods additionally use a discriminator to distinguish real images from the images generated by the decoder \citep{1804, compressnet, hific, multirealism, illm}. Diffusion-based methods usually regard the quantized latent as a condition used for a denoising diffusion process \citep{cdc, diffusion}. Unlike VAE-based methods that focus on the rate-distortion trade-off, these methods typically adopt a rate-distortion-perception trade-off during the training process \citep{rethinking}, dramatically improving the visual quality of reconstructed images.
% Generative Adversarial Networks for Extreme Learned Image Compression
% CompressNet Generative Compression at Extremely Low Bitrates
% High-Fidelity Generative Image Compression
% Multi-Realism Image Compression with a Conditional Generator
% Improving Statistical Fidelity for Neural Image Compression with Implicit Local Likelihood Models
% Lossy Image Compression with Conditional Diffusion Models
% DiffusionAutoencoders:TowardaMeaningfulandDecodableRepresentation
% Rethinking Lossy Compression The Rate-Distortion-Perception Tradeoff

The compression approaches mentioned above exhibit acceptable performance only at normal bitrates. However, at extremely low bitrates, i.e., less than 0.1 bits per pixel (bpp), VAE-based methods tend to produce severe blurriness \citep{1804}, while GAN-based methods can introduce erroneous textures \citep{sketch}. Recently, the Stable Diffusion (SD) model has been utilized as a generative prior in extremely low-bitrate image compression \citep{sketch, 2211, iclr2024}. With the powerful image priors and the generative capability of the pre-trained model, SD-based methods have demonstrated the characteristics of being less reliant on bitrates \citep{iclr2024}. However, these methods achieve extremely low bitrates by using text descriptions or sketches extracted from images to be compressed, significantly removing most of the information from the original images. Thus, the compressed bitstream cannot provide sufficient conditions for image generation and results in noticeable compression distortion. To extract essential information as effective conditions for the denoising diffusion process, we propose a novel SD-based image compression framework as shown in Fig.\:\hyperref[fig1]{1(d)}. We employ $\text{g}_\text{a}(\cdot)$ and $\text{g}_\text{s}(\cdot)$ to further compress and decompress the latent representation $z_0$, and the reverse diffusion process is conditioned on compressed latent representation $\hat{z}$ as implicit guidance to create realistic RS images.
% Text + Sketch Image Compression at Ultra Low Rates
% EXTREME GENERATIVE IMAGE COMPRESSION BY LEARNING TEXT EMBEDDING FROM DIFFUSION MODELS
% TOWARDS IMAGE COMPRESSION WITH PERFECT REALISM AT ULTRA-LOW BITRATES

Existing SD-based methods typically take full advantage of explicit conditions, such as text and sketches, which are not suitable for RS images characterized by complex and irregular ground objects. This is because the network cannot extract accurate semantic information from these conditions. For instance, when given a textual prompt like ``many houses are on both sides of the road'', the diffusion model fails to understand the precise locations of the road and houses, leading to semantically incorrect reconstructions. Therefore, to minimize semantic distortion in compressed RS images at low bitrates, it is imperative to employ explicit guidance that encapsulates spatially accurate semantic information of the original RS images.

A vector map is a graphical representation of geospatial data in which map elements such as roads and buildings are represented by geometric shapes and coordinates such as spatially positioned lines and polygons. This type of data provides semantic and structural information of ground objects and can be leveraged as semantic guidance during RS image compression and decompression. In many real-world applications, vector maps are already available or required for other purposes, such as navigation, urban planning, or geographic information systems. Utilizing these existing resources for image compression does not impose significant additional storage burdens. 
%If not, we can also store only one copy of the vector map, which requires significantly less storage space compared to rasterized RS images, on both the compression and decompression sides, and efficiently reuse it for RS images captured periodically from different satellites. 
Inspired by this, we propose an extremely low-bitrate image compression approach for RS images, named Map-Assisted Generative Compression (MAGC), which utilizes semantic information from vector maps with the pre-trained SD model to reconstruct RS images at extremely low bitrates. Specifically, we conduct the denoising diffusion process using the compressed latent representations as implicit guidance and vector maps as explicit guidance. Under implicit and explicit guidance, our MAGC can generate images with higher fidelity and better maintain semantic integrity compared to other methods. We conduct the semantic segmentation task and the experimental results indicate that our method can save 90\% bitrates than VTM-23.3 \citep{vtm} and 60\% bitrates than MS-ILLM \citep{illm} with even higher mIoU.
% Semantic Guided Large Scale Factor Remote Sensing Image Super resolution with Generative Diffusion Prior

Main contributions of this paper are as follows: 
\begin{enumerate}
\item We propose the Map-Assisted Generative Compression (MAGC) model, which integrates semantic information from vector maps into a pre-trained SD model to generate images with plausible visual quality and accurate semantic features.
\item We develop a VAE-based latent compression module (LCM) to further compress the latent representations and provide implicit guidance for the denoising diffusion process. The semantic adapter module (SAM) is employed to extract multi-scale features from vector maps as explicit guidance to generate semantically accurate images.
\item Experimental results indicate that our method outperforms others in perceptual quality and semantic accuracy. Semantic segmentation results show that our method can save 90\% bitrates than VTM-23.3 and 60\% bitrates than MS-ILLM with even better performance.  
\end{enumerate}

The rest of this paper is organized as follows. Section \ref{section2} gives a review of image compression techniques. The framework of MAGC is illustrated in Section \ref{section3}. Section \ref{section4} compares the performance of the proposed MAGC with different compression methods. In Section \ref{section5}, we conduct ablation studies and discuss the impact of each component. Finally, we summarize this paper in Section \ref{section6}.
% 引用数：26

%%%%%%%%%%%%%%%%%%%%%%%%%%%%%%%%%%%%%%%%%%%%%%%%%%%%%%%%%%%%%%%%%%%%%%%%%%%%%%%%%%%%
%%%%  Related Work (引用数：24)
\section{Related Work}\label{section2}
In this section, we present a review of image compression from the following three subfields: learned image compression models, image compression at extremely low bitrates and RS image compression.

\subsection{Learned Image Compression Models}
Over the last few years, learned image compression has made great progress and demonstrated outstanding performance. The pioneering work of Ballé et al. \citep{end-to-end} first proposed a CNN-based end-to-end learned image compression framework that uses stacked convolution layers and generalized divisive normalization (GDN) layers to achieve transform coding. Then they modified the framework as a VAE-based model and introduced a hyperprior as side information \citep{hyperprior}. Furthermore, some researchers proposed to integrate a context model with hyperprior to achieve more accurate probabilistic predictions of the latent representation \citep{channelwise,checkerboard, elic, joint}. In addition to the refinement of entropy models, multiple novel and efficient CNN-based transforms have been proposed \citep{ elic,cheng2020, tang}. Besides, inspired by the success of Vision Transformer in many vision tasks \citep{vit}, Swin-Transformer-based methods have been explored in learned image compression \citep{ stf, transformer-based}. Liu et al. \citep{mixed} proposed an efficient parallel Transformer-CNN Mixture block to improve the overall architecture of image compression models by incorporating the local modeling ability of CNN and the non-local modeling ability of transformer.
% END-TO-END OPTIMIZED IMAGE COMPRESSION
% Joint Autoregressive and Hierarchical Priors for Learned Image Compression
% Learned Image Compression with Discretized Gaussian Mixture Likelihoods and Attention Modules
% tang:Joint graph attention and asymmetric convolutional neural network for deep image compression
% vit: 
% Transformer-based transform coding

Additionally researchers have explored GAN-based and diffusion-based methods to enhance the visual quality of the reconstructed images. Based on theoretical research of the rate-perception-distortion trade-off \citep{rethinking}, many conditional GAN-based approaches \citep{hific, multirealism, illm} have been proposed. Mentzer et al. \citep{hific} proposed a novel generative image compression framework using a conditional GAN architecture. Agustsson et al. \citep{multirealism} proposed an improved decoder that navigates the distortion-realism trade-off. Muckley et al. \citep{illm} introduced an improved non-binary discriminator that is conditioned on quantized local image representations and achieve state-of-the-art perceptual quality. In contrast to these GAN-based methods, some diffusion-based methods \citep{cdc, diffusion} also show comparable performance at normal bitrates using conditional Gaussian diffusion \citep{ddpm}, but they show less stability at extremely low bitrates.
% Hoogeboom: High-fidelity image compression with score-based generative models.
% Denoising Diffusion Probabilistic Models

\subsection{Image Compression at Extremely Low Bitrates}
% Besides, Invertible Neural Network (INN) \citep{inn} and text-guided \citep{2304} methods have also been adopted in extremely low-bit compression. 
Under the extreme low-bitrate conditions, GAN-based compression methods \citep{1804, compressnet,hific,  1703, iwai, cvpr2024} have demonstrated superior performance compared to MSE-optimized VAE-based models and standard codecs. For instance, Sauturkar et al. \citep{1703} proposed a pioneering approach of generative compression utilizing a generative adversarial strategy and demonstrated impressive results. Mentzer et al. \citep{hific} incorporated rate-distortion-perceptual trade-off with a GAN-based framework, yielding reconstructions with high realism, even at 0.1bpp on the CLIC2020 test set and 0.2bpp on Kodak dataset. Jia et al. \citep{cvpr2024} introduced a Generative Latent Coding architecture that maintains high visual quality with less than 0.04 bpp on natural images and less than 0.01 bpp on facial images. Besides, text-guided transforms have also been adopted in extremely low-bitrate image compression \citep{2304}.
% Generative Compression
% Fidelity-Controllable Extreme Image Compression with Generative Adversarial Networks
% Generative Latent Coding for Ultra-Low Bitrate Image Compression
% Extremely Low Bit-rate Image Compression via Invertible Image Generation
% Multi-Modality Deep Network for Extreme Learned Image Compression

% cheng+sd,  text+sketch, iclr2024
With powerful image priors and generative capability, the pre-trained text-to-image model, such as the SD model, has been used in extremely low-bitrate image compression. For instance, Pan et al. \citep{2211} adopted learnable text embeddings and pre-trained diffusion model to generate realistic images. Lei et al. \citep{sketch} extracted textual descriptions and sketches from the original images to guide image generation. Furthermore, Careil et al. \citep{iclr2024} conditioned the diffusion model on more versatile end-to-end learned vector-quantized image features, maintaining the ability to reconstruct realistic images at ultra-low bitrate (e.g., less than 0.01 bpp).

\begin{figure*}[htb]
\centering
\includegraphics[width=\textwidth]{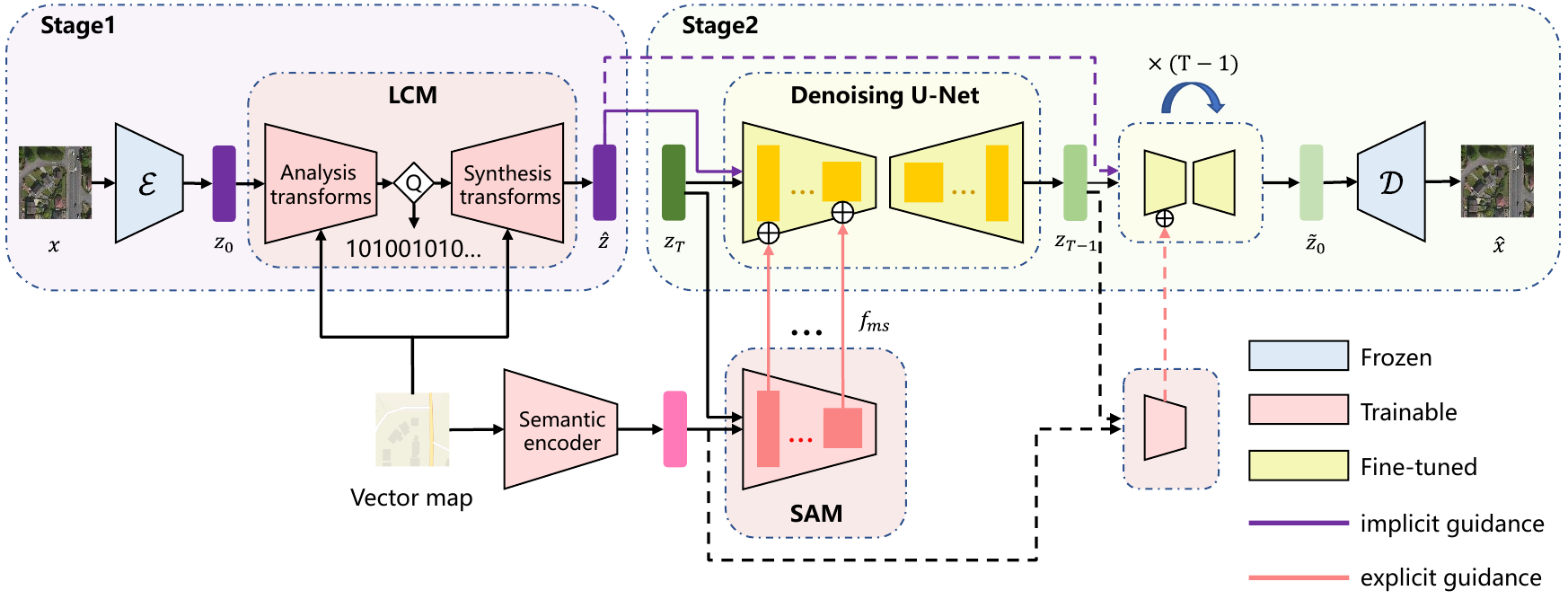}
\captionsetup{singlelinecheck=false} % 左对齐
\caption{The two-stage pipeline of the proposed MAGC. In the first stage, the latent compression module (LCM) is designed to compress the latent representation and provide implicit guidance for the conditional diffusion model. In the second stage, the semantic adapter module (SAM) is utilized to produce multi-scale features, serving as explicit guidance for the conditional diffusion process.}
\label{fig2}
\end{figure*} 

\subsection{RS Image Compression}
In the last few decades, there has been a surge in the volume of RS images, necessitating the development of effective image compression techniques. In early stages, some standard codecs based on traditional transform coding, such as JPEG and JPEG2000, were extensively applied to RS image compression \citep{JPEG2000-1, JPEG2000-2, JPEG2000-3}. These methods performed transforms from the image domain to the frequency domain, followed by quantization, and achieved significant bitrate savings. Additionally, based on the discrete wavelet transform, the Consultative Committee for Space Data Systems (CCSDS) has also been utilized before the popularity of learning-based approaches \citep{CCSDS-1, CCSDS-2}. However, prior work \citep{Efficient} has shown that severe blurriness and artifacts are prevalent in handcrafted transform coding methods, especially at low bitrates. To address this issue, some learned image compression methods have been explored for RS image compression.
% 3：JPEG2000 coding strategies for hyperspectral data. A cartoon-texture approach for JPEG/JPEG 2000 decompression based on TGV and shearlet transform. JPEG2000 encoding of remote sensing multispectral images with no-data regions
% 4: Extending the CCSDS recommendation for image data compression for remote sensing scenarios. A 13.3 Gbps 9/7M discrete wavelet transform for CCSDS 122.0-B-1 image data compression on a spacegrade SRAM FPGA,
% 5:Efficient and effective context-based convolutional entropy modeling for image compression

Encouraged by the success of VAE-based methods in natural image compression, some researchers have proposed using these models to compress RS images \citep{ tgrs2024,uplink, Oliveira, fu, Chong, xiang-long-range}. Chong et al. \citep{Chong} incorporated the Markov Random Field into the attention mechanism and showed impressive results on standard RS datasets of varying resolutions. Wang et al. \citep{uplink} used historical images as reference images for on-orbit compression based on uplink transmission procedures. Considering the reconstruction of high-frequency features such as edge information and texture characteristics, Xiang et al. \citep{tgrs2024} incorporated a dual-branch architecture based on high-frequency and low-frequency feature components. 

% Oliveira: Satellite image compression and denoising with neural networks
% Fu: Remote Sensing Image Compression Based on the Multiple Prior Information
% Chong: High-Order Markov Random Field as Attention Network for High-Resolution Remote-Sensing Image Compression
% XIANG: Remote sensing image compression with long-range convolution and improved non-local attention model

In addition to the VAE-based methods, GAN-based methods have also been developed to generate realistic RS images. Zhao et al. \citep{zhao-symmetrical} adopted several pairs of symmetrical encoder-decoder lattices and multiple discriminators to enhance edges, contours and textures in the reconstructed images. Han et al. \citep{edge} proposed an edge-guided adversarial network to simultaneously restore edge structures and generate texture details. Pan et al. \citep{tgrs2023} presented a coupled compression generation network that reconstructs image content and detailed textures separately and achieves outstanding results in the DOTA dataset at extremely low bitrates.
% Zhao: Symmetrical lattice generative adversarial network for remote sensing images compression
% Han: Edge-Guided Remote-Sensing Image Compression
% A Coupled Compression Generation Network for Remote-Sensing Images at Extremely Low Bitrates

These learning-based approaches mentioned above have shown outstanding performance in RS image compression. However, they generally focused on improving pixel-level fidelity or perceptual quality of the reconstructed images, while neglecting semantic integrity which is of paramount significance for RS images. To this end, introducing vector maps as semantic guidance in RS image compression is a promising strategy that maintains a higher degree of semantic integrity even at extremely low bitrates.
% 引用数：24

%%%%%%%%%%%%%%%%%%%%%%%%%%%%%%%%%%%%%%%%%%%%%%%%%%%%%%%%%%%%%%%%%%%%%%%%%%%%%%%%%%%%
%%%%  Proposed Method

\section{Proposed Method}\label{section3}
\subsection{Overview}
In this section, we give an overview of the proposed MAGC. As shown in Fig.\:\hyperref[fig2]{2}, our MAGC consists of two stages. In the first stage, the image is mapped into a latent representation via a pre-trained SD VAE encoder. Then the latent representation is compressed, quantized, transmitted and decompressed in the LCM. The compressed latent representation, containing the majority of the information from the original image, provides implicit guidance for the conditional diffusion model and ensures that the reconstructed image contains the same contents as the original one. The process of the first stage can be formulated as follows:
\begin{align}
    z_0&=\mathcal{E}(x), \\
    \hat{z}&=\text{LCM} (z_0, m) 
\end{align}
where $\mathcal{E}(\cdot)$ denotes the pre-trained SD VAE encoder. $x$ denotes the input image. $z_0$ and $\hat{z}$ are the original and compressed latent representation. $m$ represents the vector map.

In the second stage, we utilize the SAM to produce multi-scale features as explicit guidance, which provides additional semantic and structural information and assists in generating images with semantic accuracy. Under implicit guidance from the compressed latent representation and explicit guidance from the vector map, the conditional diffusion model is conducted in the latent space, followed by the generation of the realistic and semantically accurate reconstruction through a pre-trained SD VAE decoder. The process can be formulated as follows:
\begin{align}
f_{\text{ms}}&=\text{SAM}(\text{SE}(m), z_t), \\
\tilde{z}_0&=\text{CDM}(\hat{z},f_{\text{ms}}), \\
\hat{x}&=\mathcal{D}(\tilde{z}_0),
\end{align}
where 
$\text{SE}$ denotes the semantic encoder. 
$f_{\text{ms}}$ is the produced multi-scale features. 
$z_t$ and $\tilde{z}_0$ are the noised and denoised latent representation. $\text{CDM}$ denotes the conditional diffusion model. $\mathcal{D}(\cdot)$ denotes the pre-trained SD VAE decoder. $\hat{x}$ is the reconstructed image.

\subsection{Latent Compression Module}
The LCM is designed to further reduce the redundancy of the latent representation and provide implicit guidance for the denoising diffusion process. The network architecture of LCM is illustrated in Fig.\:\hyperref[fig3]{3}. Similar to the VAE-based learned image compression frameworks, the LCM consists of latent transform networks, hyperprior networks and a channel-wise context model. Furthermore, the spatially-adaptive denormalization (SPADE) ResBlock \citep{spade} is designed to integrate the semantic information from vector maps into the latent transform coding. The details of the LCM are presented below.

\subsubsection{VAE-based Latent Coding}
Since Ballé et al. \citep{hyperprior, end-to-end} proposed the VAE-based image compression framework, most of the learned image compression methods have followed the paradigm. Inspired by this, we model the latent compression process using VAE-based networks. The generalized latent compression framework is formulated as follows:
\begin{equation}
\begin{aligned}
    y&=\text{g}_\text{a}(z,m;\phi),   \\
    \hat{y}&=\text{Q}(y),  \\
    \hat{z}&=\text{g}_\text{s}(\hat{y},m;\theta),
\end{aligned}
\end{equation}
where $y$ denotes the compact representation obtained from $z$. $\hat{y}$ denotes the quantized counterpart of $y$. $\text{g}_\text{a}(\cdot)$ and $\text{g}_\text{s}(\cdot)$ are analysis and synthesis transforms. $\phi$ and $\theta$ are learnable parameters of the transform networks. 
$\text{Q}$ represents the quantization operation.

\begin{figure*}[htb]
\centering
\includegraphics[width=\textwidth]{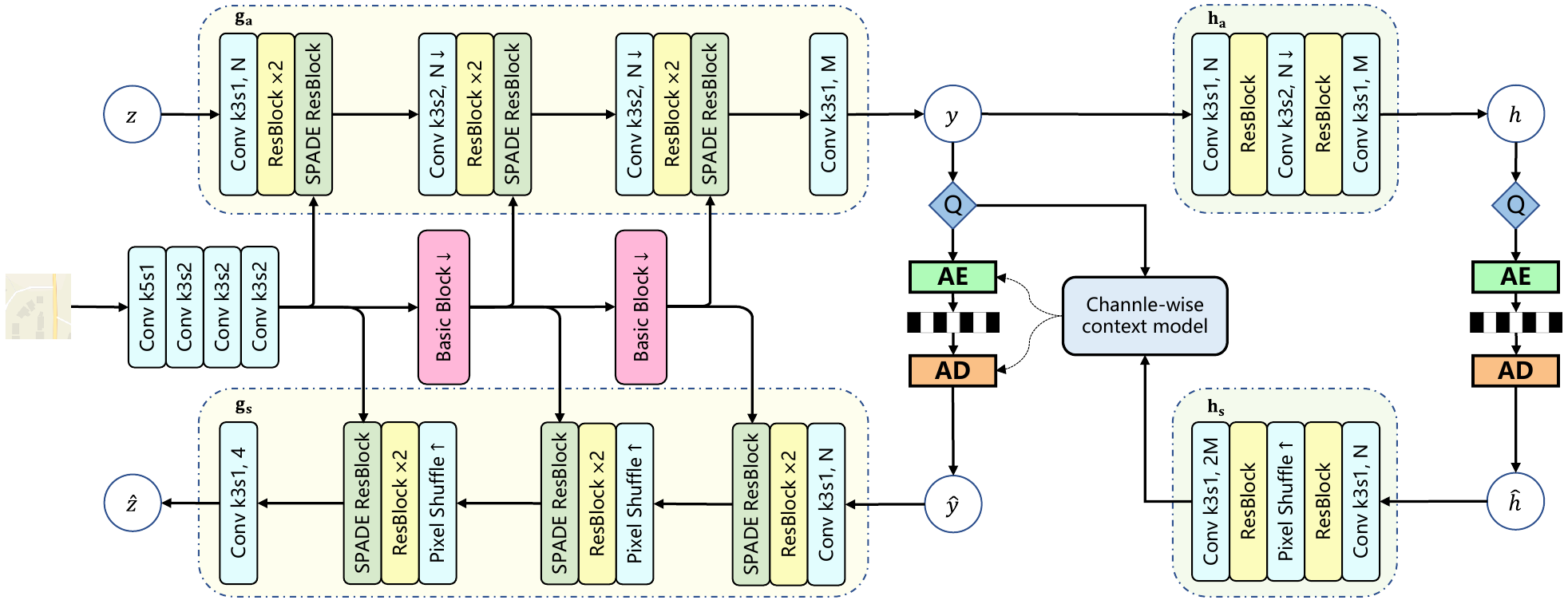}
\captionsetup{singlelinecheck=false} % 左对齐
\caption{The network architecture of the proposed LCM, consists of the latent transform networks, hyperprior networks and a channel-wise context model. \textit{Q} represents quantization. \textit{AE}, \textit{AD} represent arithmetic encoder and arithmetic decoder. $\downarrow$ and $\uparrow$ indicate downsampling and upsampling. We set \textit{N} = 128 and \textit{M} = 64 in our experiments.}
\label{fig3}
\end{figure*}

As shown in Fig.\:\hyperref[fig3]{3}, $\text{g}_\text{a}(\cdot)$ consists of several residual blocks and downsampling convolution layers, performing the transform from the input $z$ to the compact representation $y$ in the encoding process. Symmetrical to this, $\text{g}_\text{s}(\cdot)$ employs the pixelshuffle operation to perform upsampling and maps $\hat{y}$ back to $\hat{z}$. In contrast to previous methods that inject uniform noise during training to simulate the quantization error, we employ the straight-through estimator (STE) \citep{illm, ste} to pass the quantized to the decoder with backpropagation, ensuring that the decoder sees the same data distribution during training and inference.
%ste: Lossy image compression with compressive autoencoders 

Following the previous work \citep{joint}, We model each element of $\hat{y}$ as a single Gaussian distribution, and the standard deviations $\sigma$ and means $\mu$ are estimated by the entropy model. To achieve a more precise distribution estimation, the hyperprior networks \citep{hyperprior} and the channel-wise autoregressive context model \citep{channelwise} are introduced in our entropy model. As shown in the right part of Fig.\:\hyperref[fig3]{3}, the hyper encoder $\text{h}_\text{a}(\cdot)$ encodes $y$ into the side information $h$. Then $\hat{y}$ is divided into $K$ slices ($y^1,y^2,...,y^K$) alone the channel dimension, and these slices are encoded and decoded in an autoregressive manner. Specifically, estimating the distribution of the current slice $y^i$ is conditioned on both the hyperprior and the previously decoded symbols $y^{<i}$. The entire entropy model is formulated as follows:
\begin{equation}
\begin{aligned}
    h&=\text{h}_\text{a}(y;\phi_\text{h}),   \\
    \hat{h}&=\text{Q}(h),  \\
    gc&=\text{h}_\text{s}(\hat{h};\theta_\text{h}), \\
    \mu^i, \sigma^i&=\text{g}_\text{cm}(gc,y^{<i};\theta_\text{cm}), 1\leq i\leq K, \\
    p_{\hat{y}}(\hat{y}^i\vert ctx)&=[\mathcal{N}(\mu^i,(\sigma^i)^2)*\text{U}(-\frac{1}{2},\frac{1}{2})](\hat{y}^i) \\
    \text{with}\quad ctx &=(\hat{h},y^{<i},\theta_\text{h},\theta_\text{cm}),
\end{aligned}
\end{equation}
where $\text{h}_\text{a}(\cdot)$, $\text{h}_\text{s}(\cdot)$, $\phi_\text{h}$ and $\theta_\text{h}$ are hyper analysis and synthesis transform and their learnable parameters. $h$ and $\hat{h}$ are the side information and its quantized counterpart. $gc$ represents the global context, which is the output of $\text{h}_\text{s}(\cdot)$. $\text{g}_\text{cm}(\cdot)$ and $\theta_{\text{cm}}$ are the channel-wise context model and its learnable parameters.

\subsubsection{Semantic-Guided Transform}
\begin{figure}[htb]
\centering
\includegraphics[width=\linewidth]{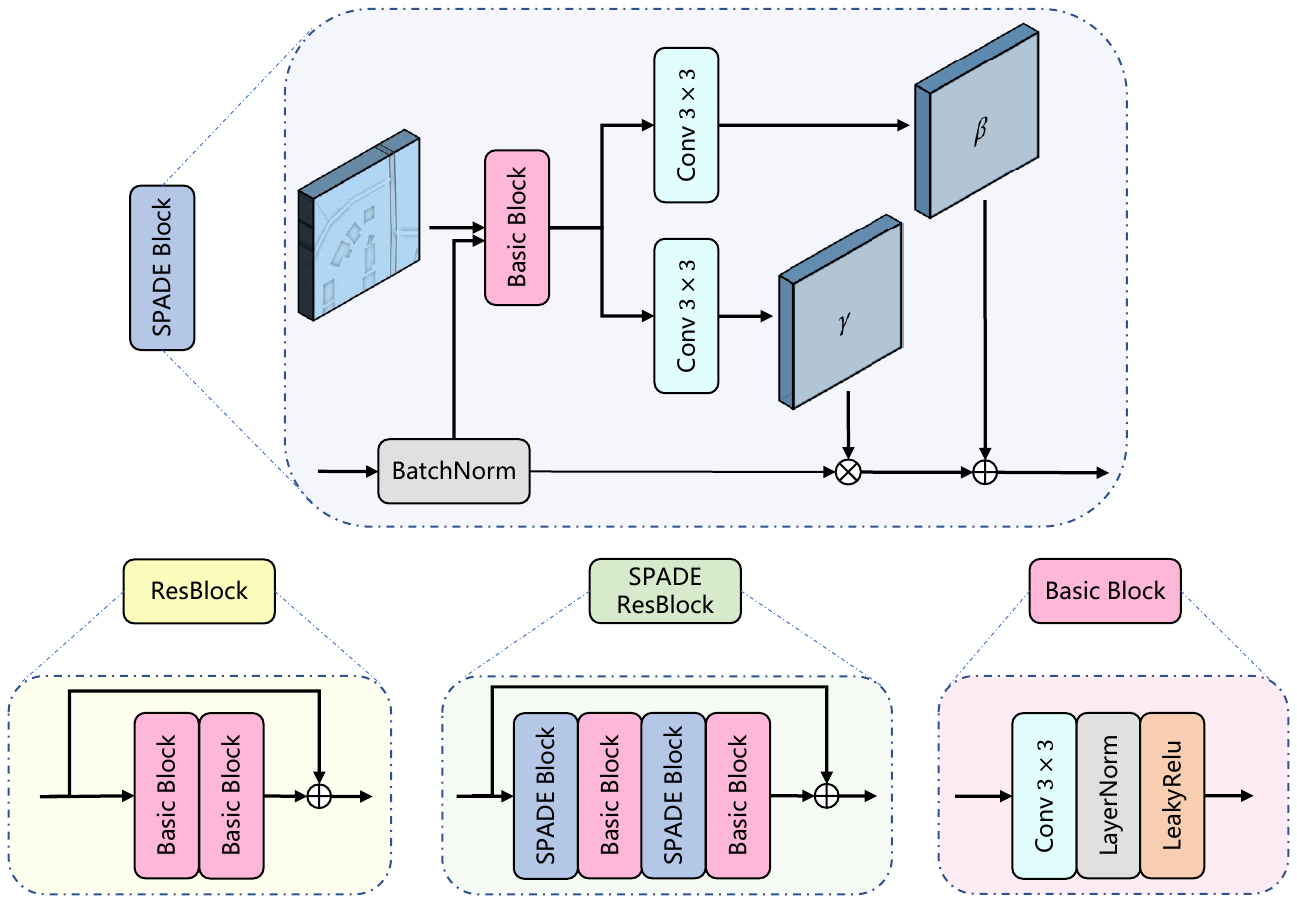}
\captionsetup{singlelinecheck=false} % 左对齐
\caption{Structures of ResBlock, SPADE ResBlock, SPADE block and basic block in our work.}
\label{fig4}
\end{figure}

As shown in Fig.\:\hyperref[fig3]{3}, we utilize the SPADE ResBlock to integrate the semantic information into the latent transform coding at each scale. The structures of the SPADE ResBlock, SPADE block and basic block are illustrated in Fig.\:\hyperref[fig4]{4}. In the SPADE block, we merge the semantic feature map and latent feature map by concatenation as the input of the first basic block. Then the merged feature map is convolved to produce the scale factor $\gamma$ and bias term $\beta$, which are multiplied and added to the normalized input element-wise. The whole process can be formulated as follows:
\begin{equation}
\begin{aligned}
    f_{\text{bn}}&=\text{BatchNorm}(f_{\text{in}}),  \\
    \gamma&=\text{Conv}_\gamma(\text{BasicBlk}([f_{\text{bn}},f_{\text{sem}}])),   \\
    \beta&=\text{Conv}_\beta(\text{BasicBlk}([f_{\text{bn}},f_{\text{sem}}])),  \\
    f_{\text{out}}&=\gamma \otimes f_{\text{bn}} + \beta,  \\
\end{aligned}
\end{equation}
where $f_{\text{in}}$, $f_{\text{bn}}$, $f_{\text{sem}}$ and $f_{\text{out}}$ are the input feature map, normalized feature map, semantic feature map and output of SPADE block. 
$\text{BasicBlk}$ denotes the basic block.
$\gamma$ and $\beta$ are the produced scale factor and bias.
$\text{Conv}_\gamma$ and $\text{Conv}_\beta$ are the convolution layers used to predict $\gamma$ and $\beta$. 
$\otimes$ represents Hadamard product.
With the help of the SPADE block, the global structural information is incorporated into the latent transform coding. In that case, the compressed latent representation is considered to be allocated a lower bitrate for storing coarse content while retaining the essential global structure. 

\subsection{Conditional Diffusion Model}
As shown in Fig.\:\hyperref[fig2]{2}, we utilize the pre-trained SD model to generate images with multiple guidance. In this section, we will introduce how SD works and how to use multiple guidance to generate realistic and semantically accurate reconstructions at extremely low bitrates.

\subsubsection{Stable Diffusion}
The classical Denoising Diffusion Probabilistic Model (DDPM) performs adding and removing noise in pixel space \citep{ddpm}, which is time-consuming for both training and inference. In contrast, the SD model conducts the diffusion process in the latent space with a much smaller spatial resolution \citep{ldm}. Generally, it consists of a pre-trained VAE and a corresponding U-Net denoiser. Firstly, the VAE is trained to convert images into latent space and then reconstruct them, which is formulated as follows:
% High-Resolution Image Synthesis with Latent Diffusion Models

\begin{equation}
\begin{aligned}
    z_0&=\mathcal{E}(x),  \\
    \hat{x}&=\mathcal{D}(z_0),   \\
\end{aligned}
\end{equation}
where $x$ and $\hat{x}$ are the input and reconstructed image. 
$z_0$ denotes the latent representation. 
For the latent forward diffusion process, $z_0$ is progressively corrupted by adding Gaussian noise through a Markov chain, which can be described as:
\begin{equation}
\begin{aligned}
    z_t&=\sqrt{\bar{\alpha}_t}z_0+\sqrt{1-\bar{\alpha}_t}\epsilon, \quad \epsilon \sim \mathcal{N}(0, \textbf{I}),
\end{aligned}
\end{equation}
where $z_t$ is the noised latent representation at step $t$. 
$\bar{\alpha}_t$ is a pre-set hyperparameter controlling the intensity of the added noise. 
Then a denoiser function, $\epsilon_\theta$, is trained to estimate the noise injected to $z_t$. The optimization process can be formulated as follows:
\begin{equation}
\begin{aligned}
    \mathcal{L}=\mathbb{E}_{z_0,\epsilon,t}\left\|\epsilon- \epsilon_\theta(z_t,t)  \right\|^2,
\end{aligned}
\end{equation}
where $t$ is the timestep sampled from [0, $T$].
During inference, the input noised latent $z_T$ is randomly sampled from a standard Gaussian distribution. After $T$ iterations of the backward diffusion processes, the denoised latent is mapped back to pixel space by the pre-trained SD VAE decoder $\mathcal{D}(\cdot)$.

\subsubsection{Denoising with Multiple Guidance}
In our work, the denoising diffusion process is conditioned on both implicit and explicit guidance. Specifically, $\hat{z}$, the output of LCM, contains most of the details from the original images, serving as the implicit guidance, while the vector maps provide additional semantic and structural information as explicit guidance. 

For implicit guidance, we concatenate $\hat{z}$ with $z_t$ as the input of the denoising U-Net at each time step. To adapt the modified dimension, we extend the first convolutional layer by increasing the number of channels that corresponds to the channel dimension of $\hat{z}$, i.e. from 4 to 8. The additional channels are initialized to zero during training. Since $\hat{z}$ contains the compressed information from $z_0$, the final output of the denoising process is controlled toward the original latent representation with implicit guidance.

\begin{figure}[htb]
\centering
\includegraphics[width=\linewidth]{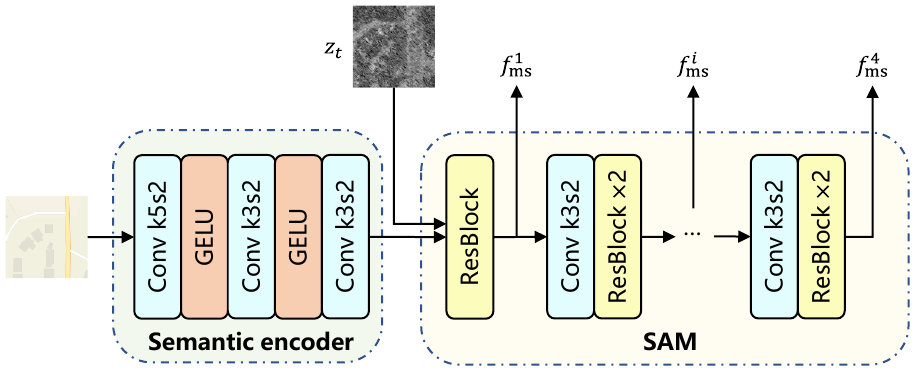}
\captionsetup{singlelinecheck=false} % 左对齐
\caption{Network architecture of semantic encoder and semantic adapter module (SAM).}
\label{fig5}
\end{figure}

For explicit guidance, we use a semantic encoder to transform the vector map into a semantic feature map, which is then transformed into multi-scale conditional features $f_{\text{ms}}$ in SAM. The network architectures of the semantic encoder and SAM are shown in Fig.\:\hyperref[fig5]{5}. Similar to previous work \citep{adapter}, the vector map is first downsampled to match the dimension of the noised latent $z_t$, and then concatenated with $z_t$ as input to the first residual block, yielding the feature $f_{\text{ms}}^1$. In each subsequent scale, a convolutional layer and two residual blocks are employed to extract corresponding feature $f_{\text{ms}}^i$. Note that the produced multi-scale features $f_{\text{ms}}=\{f_{\text{ms}}^1,f_{\text{ms}}^2,f_{\text{ms}}^3,f_{\text{ms}}^4\}$ have the same dimension with the intermediate features $f_{\text{enc}}=\{f_{\text{enc}}^1,f_{\text{enc}}^2,f_{\text{enc}}^3,f_{\text{enc}}^4\}$ in the encoder side of the U-Net denoiser. These conditional features $f_{\text{ms}}$ are then added with $f_{\text{enc}}$ at each scale. By integrating the explicit guidance with the denoising diffusion process, the reconstructions are considered to be semantically consistent with the original images even at extremely low bitrates.

% T2I-Adapter: Learning Adapters to Dig out More Controllable Ability for Text-to-Image Diffusion Models
% DiffBIR Towards Blind Image Restoration with Generative Diffusion Prior

\subsection{Training Strategy} The overall optimization objectives consist of three trade-off terms: rate loss $\mathcal{L}_{\text{rate}}$, latent distortion loss $\mathcal{L}_{\text{ld}}$ and diffusion loss $\mathcal{L}_{\text{diff}}$. The loss function can be formulated as:
\begin{equation} 
\begin{aligned}
    \mathcal{L}=\lambda_{\text{rate}}\mathcal{L}_{\text{rate}} + \lambda_{\text{ld}}\mathcal{L}_{\text{ld}} + \lambda_{\text{diff}}\mathcal{L}_{\text{diff}},
\end{aligned}
\label{eq12}
\end{equation}
where $\lambda_{\text{rate}}$, $\lambda_{\text{ld}}$ and $\lambda_{\text{diff}}$ are the corresponding weights. However, we have found that the joint optimization described in \:\hyperref[eq12]{(12)} leads to instability of training and difficulty in convergence. In contrast, we propose a two-stage training strategy for the optimization of the latent compression module and conditional diffusion model respectively. 

In the first stage, we introduce the rate-distortion trade-off to optimize the LCM, which is described as:
\begin{align}
    \mathcal{L}_{\text{1st}}&=\mathcal{L}_{\text{rate}} + \lambda\mathcal{L}_{\text{ld}}, \nonumber\\
    \mathcal{L}_{\text{rate}}&=\mathbb{E}_{z_0\sim p_z}[-{\rm log_2}\mbox{\,}p_{\hat{y}}(\hat{y})]+\mathbb{E}_{z_0\sim p_z}[-{\rm log_2}\mbox{\,}p_{\hat{h}}(\hat{h})], \nonumber\\
    \mathcal{L}_{\text{ld}}&=\left\|\hat{z}-z_0 \right\|^2,
\label{eq13}
\end{align} 
where $\lambda$ is the Lagrange multiplier that controls the expected trade-off between rate and latent distortion.
$p_z$ is the distribution of latent representations obtained from input source images. 
$p_{\hat{y}}$ and $p_{\hat{h}}$ are the probability distribution of $\hat{y}$ and $\hat{h}$ for the entropy coding.

In the second stage, we finetune the decoding transform $\text{g}_\text{s}(\cdot)$ and $\text{h}_\text{s}(\cdot)$ of LCM to condition the denoising diffusion process adaptively. Since the SD model is pre-trained on natural images, we finetune the linear layers on our training set to improve the compression performance. The trainable parameters are illustrated in Fig.\:\hyperref[fig2]{2} and the loss function is formulated as follows:
\begin{equation} 
\begin{aligned}
    \mathcal{L}_{\text{2nd}}&=\mathbb{E}_{z_0,c,\epsilon,t}\left\|\epsilon- \epsilon_\theta(z_t,c,t)  \right\|^2 \\
    \text{with}\quad c &=(\hat{z},f_{\text{ms}}),
\end{aligned}
\end{equation}
where $c$ presents the conditional control that contains implicit and explicit guidance.

%%%%%%%%%%%%%%%%%%%%%%%%%%%%%%%%%%%%%%%%%%%%%%%%%%%%%%%%%%%%%%%%%%%%%%%%%%%%%%%%%%%%
%%%%  Experiments
\section{Experiments}\label{section4}
\subsection{Experimental Settings}
\subsubsection{Datasets}

In this paper, we collect paired RS images and vector maps from different regions as training and testing datasets. All RS images are sourced from the World Imagery, while all vector maps are acquired from the Open Street Map. The training set comprises 45,000 image pairs, while the test set contains 4,500 image pairs. The resolution of each image is $256\times256$, and the spatial resolution is 1.07 m/pixel. The dataset will be publicly available to the community to facilitate the advancements on RS image compression.

% 这里可以多介绍一点
\subsubsection{Baseline Methods}
To demonstrate the effectiveness of the proposed method, we compare it with standard codecs and state-of-the-art learning-based methods, including BPG \citep{bpg}, VTM-23.3 \citep{vtm}, HiFiC \citep{hific}, MS-ILLM \citep{illm}, STF \citep{stf}, ELIC \citep{elic} and HL-RSCompNet \citep{tgrs2024}. Among them, BPG\footnote{\url{https://bellard.org/bpg/}} and VTM-23.3\footnote{\url{https://vcgit.hhi.fraunhofer.de/jvet/VVCSoftware\_VTM}} are image compression standards. HiFiC\footnote{\url{https://github.com/Justin-Tan/high-fidelity-generative-compression}} and MS-ILLM\footnote{\url{https://github.com/facebookresearch/NeuralCompression/tree/main/projects/illm}} are classic and improved GAN-based approaches. STF\footnote{\url{https://github.com/Googolxx/STF}} and ELIC\footnote{\url{https://github.com/VincentChandelier/ELiC-ReImplemetation}} are MSE-optimized VAE-based frameworks. HL-RSCompNet\footnote{\url{https://github.com/shao15xiang/HL-RSCompNet}} is the state-of-the-art learning-based method for RS image compression. All the learning-based baseline methods are retrained and tested on the same dataset for fair comparisons.

\subsubsection{Implementation Details}
Our MAGC is built based on the CompressAI\footnote{\url{https://github.com/InterDigitalInc/CompressAI/tree/master/compressai}} \citep{compressai} and the Stable Diffusion 2.1-base\footnote{\url{https://github.com/Stability-AI/stablediffusion}} model \citep{ldm}. We implement the whole model using the PyTorch framework and all experiments are conducted on three NVIDIA GeForce RTX 3090 GPUs.
% Compressai: a PyTorch library and evaluation platform for end-to-end compression research

We train the MAGC in two-stages, we first optimize the LCM with a batch size of 16 for 250K iterations. We set $\lambda$ = \{0.10, 0.20, 0.39, 0.67, 0.91, 1.25\} to control the compression ratio. Secondly, we optimize the decoder of LCM, semantic encoder, SAM and the linear layers of denoising U-net, which is around 15\% of all weights. We adopt a grid of 1,000 timesteps to finetune the diffusion model. The batch size is set to 48 and the entire network converges after another 150K iterations. For both training stages, the AdamW optimizer is adopted with a learning rate of $5\times10^{-5}$, and a linear warmup is applied for the first 10k training iterations. At inference time, we employed DDPM sampling with 50 denoising steps to create images.

\subsubsection{Evaluation Metrics}
To evaluate compression performance, both full-reference and no-reference image quality assessment (IQA) metrics are adopted. For full-reference metrics, learned perceptual image patch similarity (LPIPS) \citep{lpips} and deep image structure and texture similarity (DISTS) \citep{dists} are developed that focus on preservation of texture and human perception. For no-reference metrics, we utilize multi-scale image quality transformer (MUSIQ) \citep{musiq} to capture image quality at different granularities and employ fréchet inception distance (FID) \citep{fid} to measure the statistical fidelity of the reconstructed images. Besides, we comprehensively compare codecs with different bitrates using Bjontegaard (BD) metrics \citep{bd}: BD-LPIPS, BD-DISTS and BD-FID, which can be regarded as the average improvement between codecs under the same bitrate. We also report BD-PSNR and BD-SSIM, although these distortion metrics are less meaningful on extremely low-bitrate image compression problems \citep{sketch,iclr2024}.
% The Unreasonable Effectiveness of Deep Features as a Perceptual Metric
% Image Quality Assessment: Unifying Structure and Texture Similarity
% Musiq: Multiscale image quality transformer
% GANs Trained by a Two Time-Scale Update Rule Converge to a Local Nash Equilibrium
% Calculation of average psnr differences between rd-curves

Unlike other metrics calculated from a pair of images or a single image, FID assesses the similarity between the distributions of original and reconstructed images. To more accurately describe the similarity, we follow \citep{hific} and extract patches to calculate FID. For each $H\times W$ image, we first extract $\lfloor H/f \rfloor \cdot \lfloor W/f \rfloor$ non-overlapping $f\times f$ patches. Then we shift the extraction windows by $f/2$ in both dimensions to get another $(\lfloor H/f \rfloor -1)\cdot(\lfloor W/f \rfloor -1)$ patches. In our experiments, $f$ is set to 128, and we can get 22,500 patches on the test set for subsequent calculations.

\subsection{Comparison with State-of-the-Art Methods}
\subsubsection{Quantitative Comparisons}
\begin{figure*}[htb]
\centering
\includegraphics[width=\textwidth]{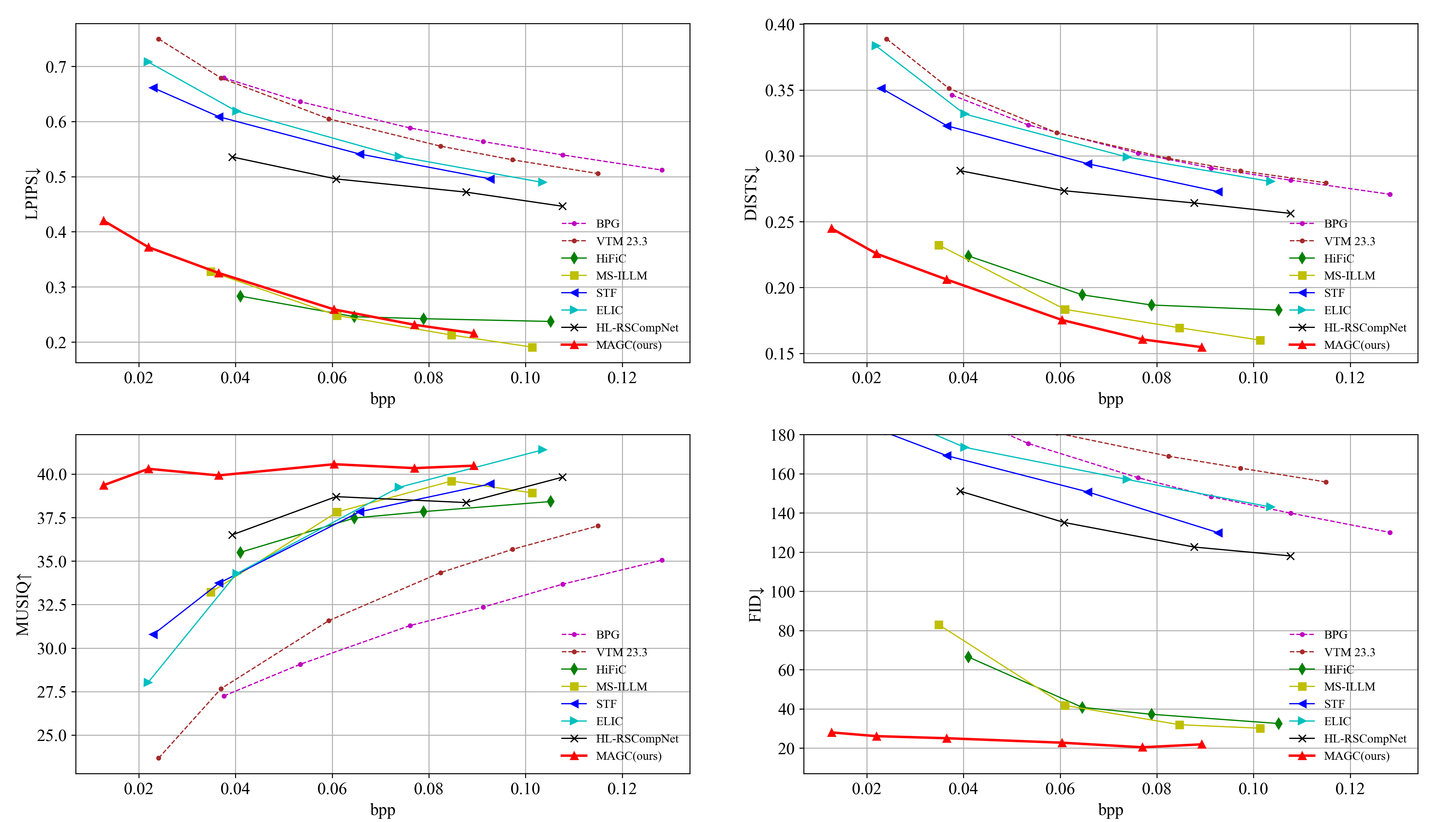}
\captionsetup{singlelinecheck=false} % 左对齐
\caption{Quantitative comparisons with state-of-the-art approaches across various perception metrics (LPIPS$\downarrow$/ DISTS$\downarrow$/ MUSIQ$\uparrow$/ FID$\downarrow$) on the test set.}
\label{fig6} 
\end{figure*}

Rate-perception curves produced by our MAGC and baseline methods are shown in Fig.\:\hyperref[fig6]{6}. It is obvious that our MAGC outperforms baseline methods in terms of perceptual quality at extremely low bitrates in terms of various perception metrics. For LPIPS, our method is comparable with GAN-based HiFiC and MS-ILLM, which are optimized based on LPIPS loss. For DISTS, MUSIQ and FID, our method outperforms all baseline methods, indicating that our method has the best potential to enhance human perceptual quality which is conducive to visual interpretation.

The comparison results of BD-metrics (Anchor: VTM-23.3) are shown in Table \ref{table1}. We calculate the quality improvement compared to the anchor over a certain bitrate range and the best results are highlighted in bold. As shown in Table \ref{table1}, our MAGC outperforms all baseline methods in terms of BD-LPIPS, BD-DISTS and BD-FID. These results further demonstrate that the pre-trained SD model can be used to generate perceptually pleasing and high-realism images with implicit and explicit guidance, even at extremely low bitrates. We also note that distortion metrics like BD-PSNR and BD-SSIM are not favorable to our method. This is because the compressed images are generated using pre-trained diffusion model as image prior, not considering pixel-level distortion during optimization \citep{sketch, iclr2024}. Other generative compression approaches like HiFiC and MS-ILLM optimized by rate-distortion-perceptual trade-off show better pixel-level fidelity than our MAGC, but the images generated by them are inferior to our method in terms of perceptual quality, which should be paid more attention in such extreme scenarios.

\begin{table}[htb]
\setlength\tabcolsep{0.9pt}
\renewcommand{\arraystretch}{1.3}
\caption{BD-metrics calculated by different methods on the test set. VTM-23.3 is the anchor.}
\label{table1}
\resizebox{0.48\textwidth}{!}{
\begin{tabular}{ l  c  c c c c}
\toprule
Methods & BD-LPIPS$\downarrow$ & BD-DISTS$\downarrow$ & BD-FID$\downarrow$ & BD-PSNR$\uparrow$ & BD-SSIM$\uparrow$\\
\hline 
\multicolumn{3}{l}{\textit{Standard codecs}} \\
VTM-23.3 & 0.0000 & 0.0000 & 0.0000 & 0.0000 & 0.0000\\
BPG & 0.0172 & -0.0013 & -11.965 & -0.2351 & -0.0101\\
\hline 
\multicolumn{3}{l}{\textit{MSE Baselines}} \\
% \textit{MSE Baselines}\\
STF & -0.0628 & -0.0249 & -28.861 & 0.5599 & 0.0289\\
ELIC & -0.0429 & -0.0100 & -19.523 & 0.3701 & 0.0158\\
HL-RSCompNet & -0.1424 & -0.0574 & -39.661 & \textbf{1.0293} & \textbf{0.0587}\\
\hline 
\multicolumn{3}{l}{\textit{Generative Compression Methods}} \\
% \textit{Genarative Compression}\\
HiFiC & -0.3394 & -0.1152 & -133.56 & -2.0506 & -0.0505\\
MS-ILLM & -0.3513 & -0.1302 & -133.05 & -1.7533 & -0.0432\\
MAGC (ours) & \textbf{-0.3521} & \textbf{-0.1464} & \textbf{-166.06} & -2.9213 & -0.1165\\
\bottomrule
\end{tabular}
}
\end{table}

\subsubsection{Qualitative Comparisons}
\begin{figure*}[htb]
\centering
\includegraphics[width=0.9\textwidth]{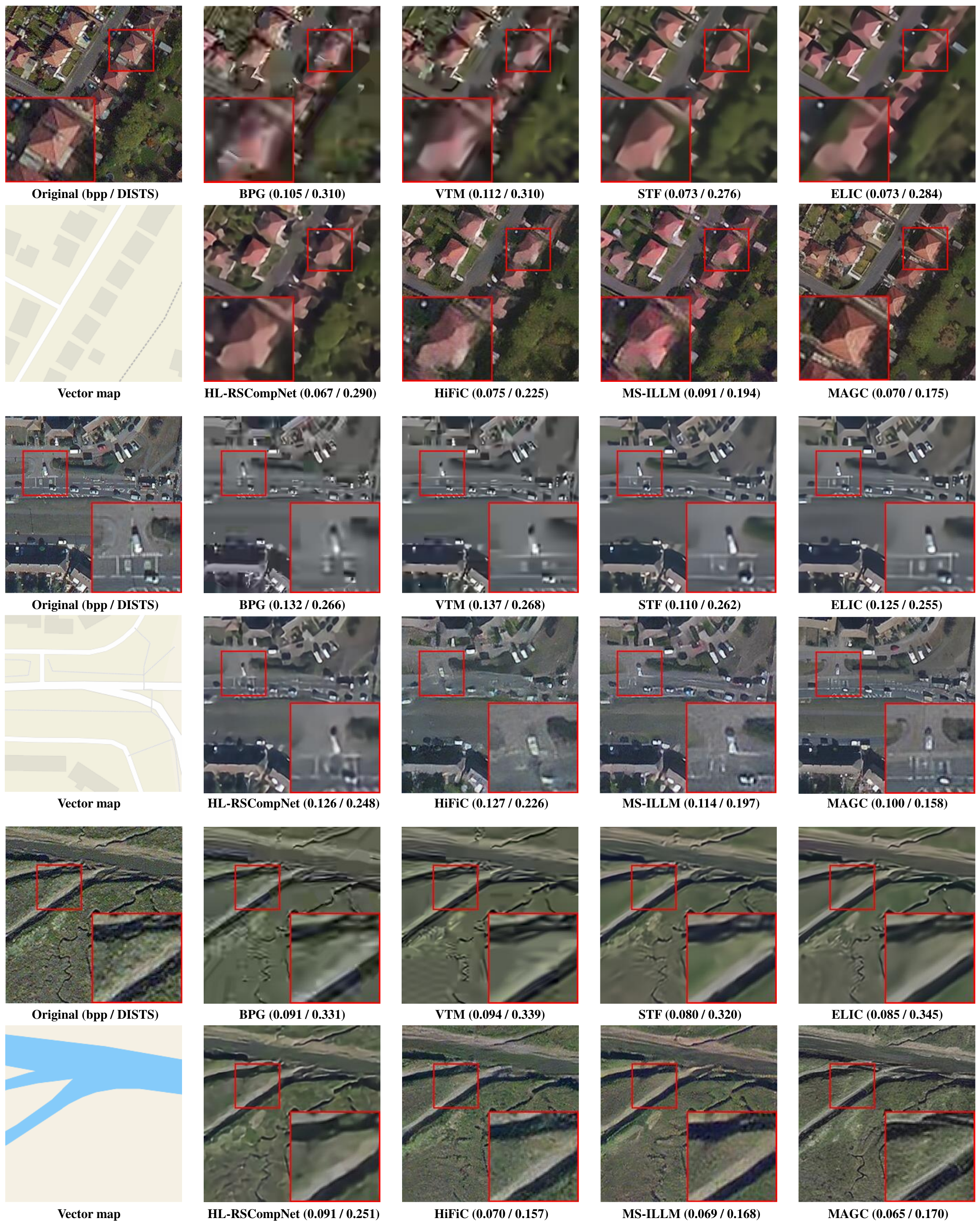}
\captionsetup{singlelinecheck=false} % 左对齐
\caption{Qualitative examples produced by various methods. These three examples are characterized by houses, roads, and rivers respectively. We employ compressed images with the closest bitrates for fair comparison.}
\label{fig7} 
\end{figure*}

To demonstrate the effectiveness of the proposed method on visual perception, we select three examples characterized by different ground objects and visualize the output images of various methods in Fig.\:\hyperref[fig7]{7}. It is easy to observe that the standard codecs BPG and VTM are more likely to produce severe artifacts at low bitrates. The reconstructed images of STF, ELIC and HL-RSCompNet can better preserve the structure information from the original images due to the VAE-based architectures and MSE loss function, but they still suffer from blurriness. GAN-based HiFiC and MS-ILLM pay more attention to human judgment using the rate-distortion-perception trade-off to produce sharper edges and more texture features. However, some complex textures, like the shape of the houses and roads,  are difficult to reconstruct accurately. In contrast to these methods, we utilize vector maps to provide semantic and structural information so that our MAGC can generate realistic and semantically accurate reconstructions at extremely low bitrates.

\subsection{Effect on Semantic Segmentation} In addition to objective fidelity and visual quality, another requirement of RS image compression lies in maintaining their performance on relevant downstream vision tasks, such as object detection, scene recognition, semantic segmentation and others. To this end, we additionally evaluate the proposed MAGC and baseline methods on the semantic segmentation task and demonstrate the superior practicality of our method.

\begin{figure*}[htb]
\centering
\includegraphics[width=\textwidth]{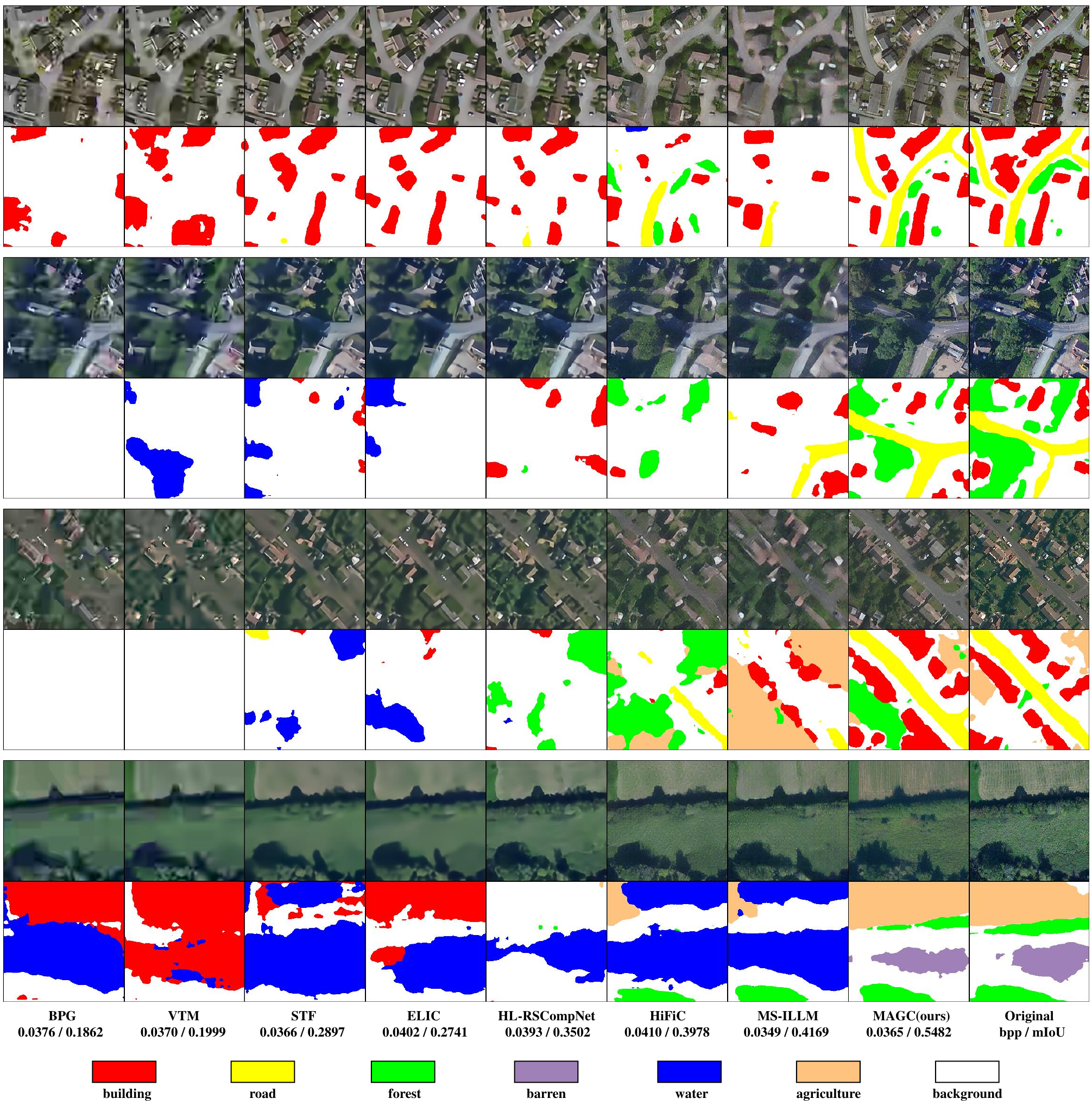}
\captionsetup{singlelinecheck=false} % 左对齐
\caption{Visual comparison of compressed images and corresponding semantic segmentation results from various methods. Values at the very bottom are the averages measured on the test set.}
\label{fig8} 
\end{figure*} 

\begin{figure}[htb]
\centering
\includegraphics[width=\linewidth]{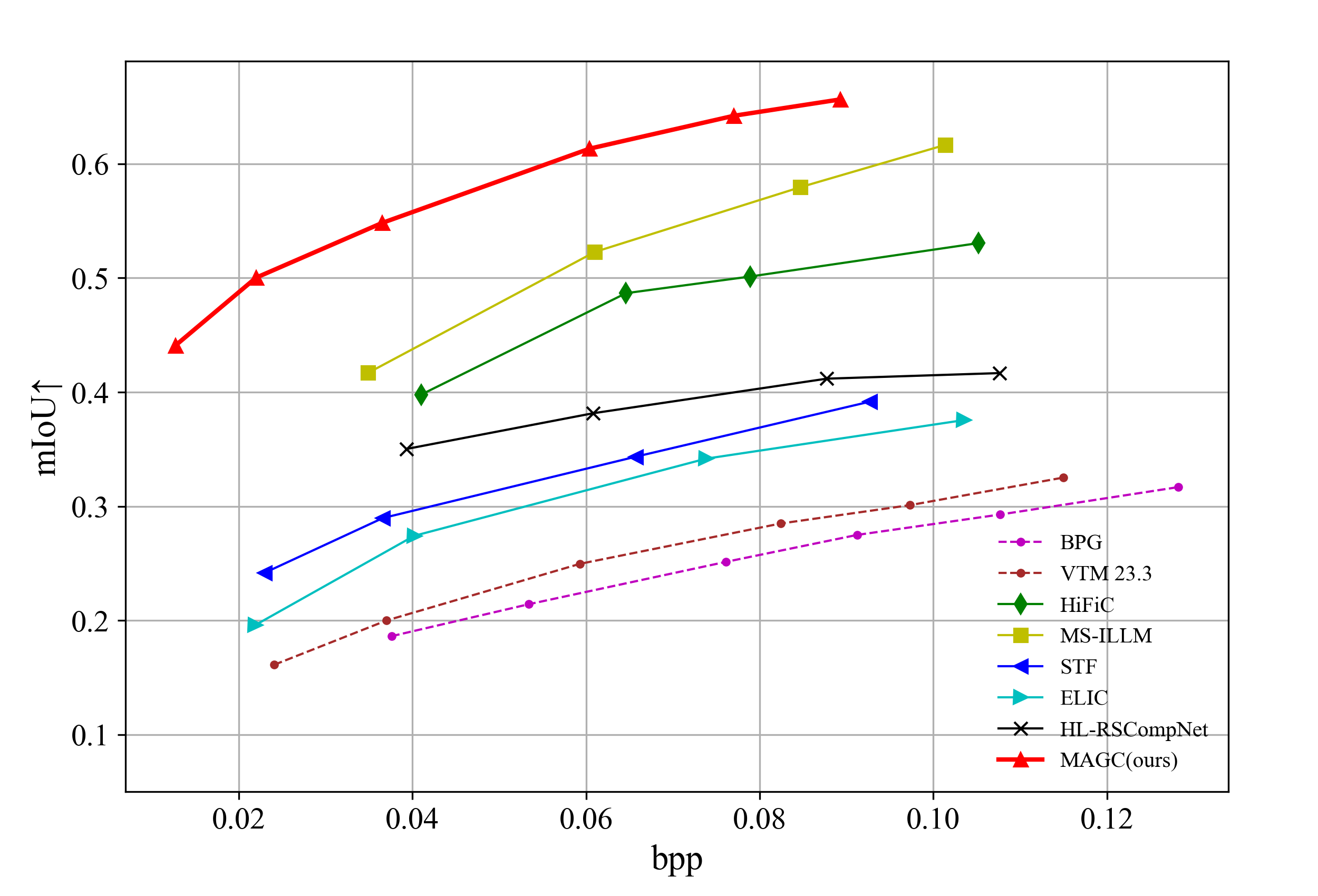}
\captionsetup{singlelinecheck=false} % 左对齐
\caption{Quantitative comparisons of mIoU calculated by various methods.}
\label{fig9} 
\end{figure} 

\begin{figure}[htb]
\centering
\includegraphics[width=\linewidth]{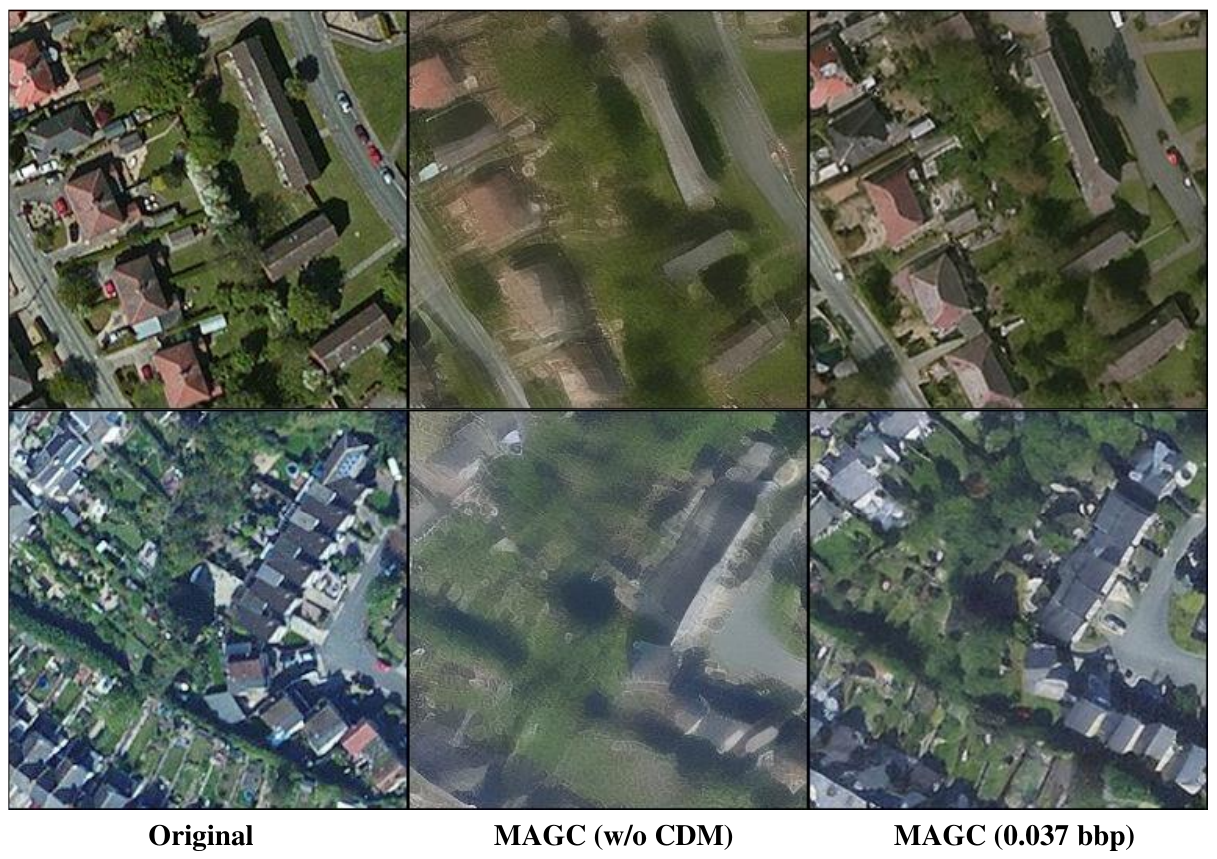}
\captionsetup{singlelinecheck=false} % 左对齐
\caption{Visual examples of the ablation study on the conditional diffusion model.}
\label{fig10} 
\end{figure} 

Note that the ground truth of semantic segmentation is unavailable, we leverage the relative performance toward the results of original images to assess the reconstructions of different approaches. In our experiments, the RS foundation model RVSA \citep{rvsa} finetuned on the LoveDA dataset \citep{loveda} is used to segment the images. Some examples of the compressed images and corresponding semantic segmentation results are shown in Fig.\:\hyperref[fig8]{8}. The visual results show that images produced by our method are semantically closer to the original images than other methods at similar bitrates. For RS images with irregular ground objects, different objects reconstructed by other methods tend to blend together, making it difficult for segmentation. In contrast, with the help of vector maps, our method can accurately distinguish the edges and generate clear textures, resulting in better segmentation outcomes. For instance, the houses and roads generated by our method are more accurate in terms of locations and shapes.
% Advancing plain vision transformer toward remote sensing foundation model.
% Loveda: A remote sensing land-cover dataset for domain adaptive semantic segmentation

We also employ the mean intersection over union (mIoU) to measure the semantic similarity between reconstructed and original images quantitatively. As shown in Fig.\:\hyperref[fig9]{9}, our MAGC outperforms all baseline methods in terms of mIoU by integrating semantic guidance from vector maps with pre-trained SD model. By analyzing the quantitative results across various bitrates, we find that our MAGC achieves a relatively high mIoU ($\textgreater$0.4) at a bitrate less than 0.02, saving 90\% rate than VTM-23.3 and 60\% rate than MS-ILLM with even better performance.

%%%%%%%%%%%%%%%%%%%%%%%%%%%%%%%%%%%%%%%%%%%%%%%%%%%%%%%%%%%%%%%%%%%%%%%%%%%%%%%%%%%%
%%%%  Ablation Studies and Discussions
\section{Ablation Studies and Discussions}\label{section5}
To better analyze and understand the proposed method, we conduct ablation studies on each component in this section.

\begin{table}[htb]
\setlength\tabcolsep{5pt}
\renewcommand{\arraystretch}{1.3}
\caption{Ablation study on the pre-trained Stable Diffusion model.}
\label{table2}
\resizebox{0.48\textwidth}{!}{
    \begin{tabular}{ l  c  c c c c}
    \toprule
    Methods & PSNR$\uparrow$ & LPIPS$\downarrow$ & DISTS$\downarrow$ & FID$\downarrow$ & mIoU$\uparrow$\\
    \hline 
    Pre-trained SD VAE & 26.005 & 0.1369 & 0.0964 & 11.210 & 0.7566\\
    % Autoencoder + CDM & 26.2001 & 0.1112 &0.0848 & 5.9511 & 0.8064\\
    \hline 
    MAGC (w/o CDM)  & 22.334 & 0.5867 & 0.3605 & 127.71 & 0.2798\\
    MAGC (0.037 bpp) & 21.763 & 0.3253 &0.2061 & 25.029 & 0.5482\\
    \bottomrule
    \end{tabular}
}
\end{table}

\subsection{Impact of Pre-Trained Stable Diffusion Model}
We first evaluate the impact of the pre-trained SD model, which consists of a VAE and a latent diffusion model. To explore the information lost in the transforms between pixel space and latent space, we only employ the pre-trained SD VAE to compress and reconstruct images. The results on various metrics are shown in Table \ref{table2}, which can be regarded as the upper bound of the performance of SD-based methods. 
% To assess the generative capability of the conditional diffusion model (CDM), we individually finetuned the linear layers of the diffusion model conditioned on the latent variables, and the inference results are shown in Table \ref{table2}. We find that the results of ``Autoencoder'' and ``Autoencoder + CDM'' is quantitatively close to each other, indicating that stable diffusion model has the potential to generate high-fidelity and high-realism images under implicit guidance.

To further demonstrate the powerful image priors and generative capability of the pre-trained diffusion model, we perform the proposed LCM alone, bypassing the conditional diffusion model as a prior and using the compressed latent representation as the input of the pre-trained SD VAE decoder. As shown in ``MAGC w/o CDM'' of Table \ref{table2}, the reconstructed images show significantly lower perceptual quality due to the absence of the pre-trained diffusion model. The reason lies in the fact that VAE-based LCM can only retain vague outlines at extremely low bitrates, while clear texture structures are generated by the pre-trained diffusion model. We show visual examples of this configuration in Fig.\:\hyperref[fig10]{10}.

\subsection{Impact of Semantic Guidance} 
\begin{figure}[htb]
\centering
\includegraphics[width=\linewidth]{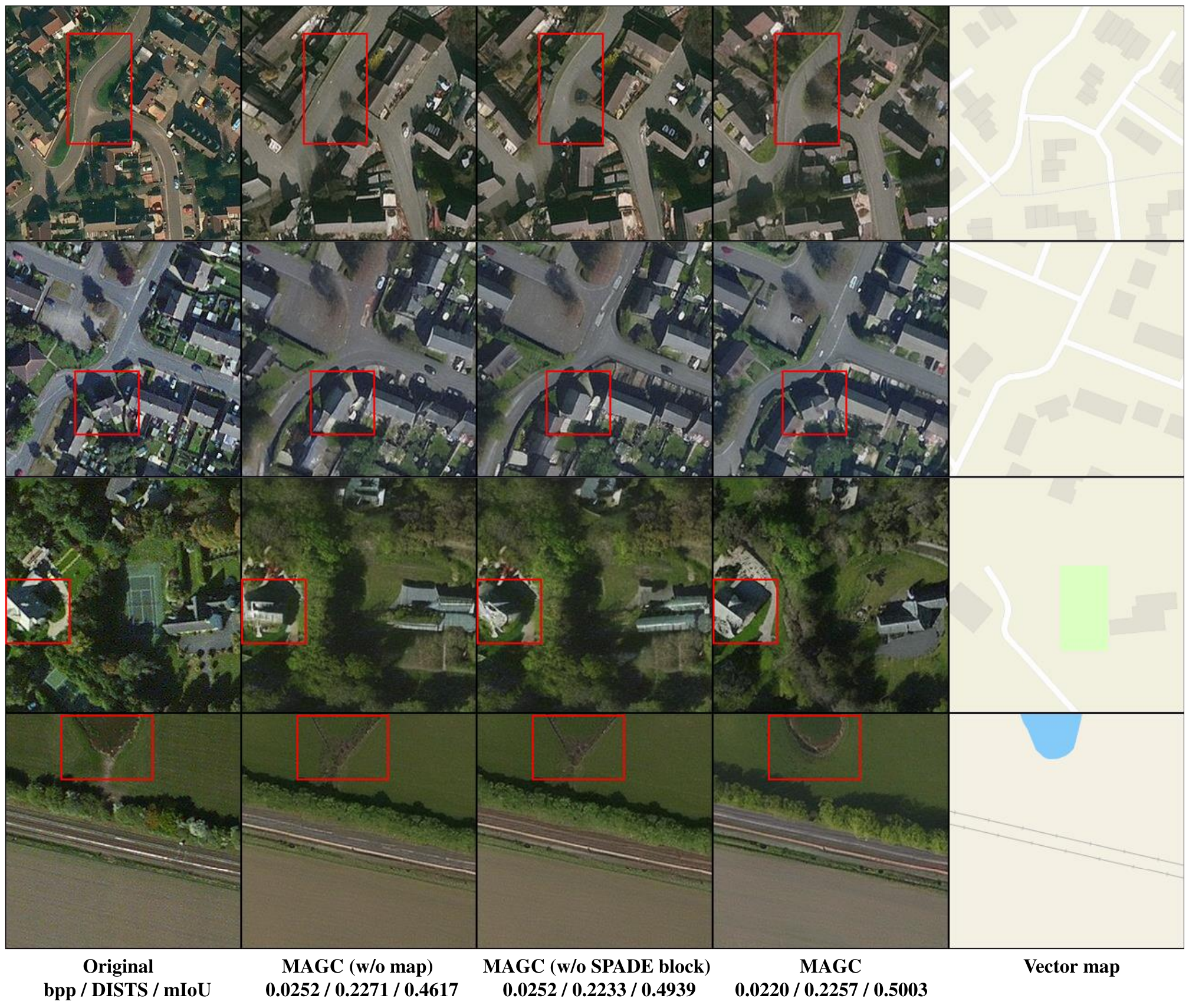}
\captionsetup{singlelinecheck=false} % 左对齐
\caption{Visual examples of the ablation study on the semantic guidance in both stages. Values at the very bottom are the averages measured on the test set.}
\label{fig11} 
\end{figure}

\begin{figure*}[htb]
\centering
\includegraphics[width=\textwidth]{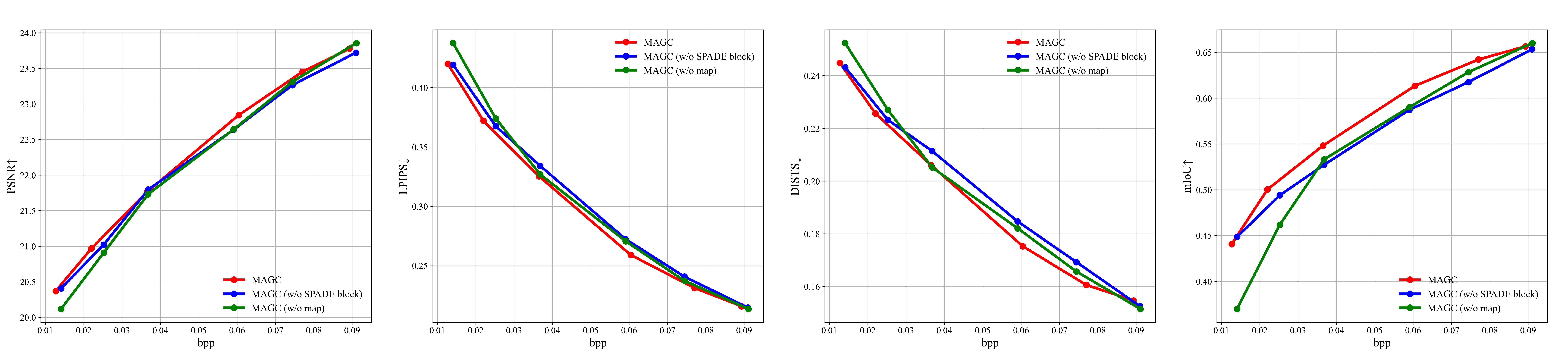}
\captionsetup{singlelinecheck=false} % 左对齐
\caption{Ablation study on the introduction of vector maps, which serve as semantic guidance in both stages of our MAGC.}
\label{fig12} 
\end{figure*}

We inject semantic information from vector maps in both stages by designing the SPADE block and SAM. To explore the contribution of vector maps, we remove these modules and test the compression performance. In addition, we also consider the scenarios where vector maps are only used on the decoding side, i.e., we eliminate SPADE blocks and retain SAM to provide explicit guidance for denoising diffusion process. The metrics evaluated by different configurations are shown in Fig.\:\hyperref[fig12]{12}. In general, removing vector maps results in much worse perceptual quality and segmentation results at relatively low bitrates, while the results using vector maps only on the decoding side are slightly inferior to the full MAGC. On the other hand, these three configurations achieve similar compression performance at higher bitrates. We also report BD-metrics of these configurations in Table \ref{table3}, indicating that our MAGC makes full use of vector maps to produce perceptually pleasing and semantically accurate images. We present visual comparisons in Fig.\:\hyperref[fig11]{11}. We notice that MAGC (w/o map) struggles to distinguish complex ground objects in such extreme scenarios, leading to erroneous reconstructions such as houses and roads. With the help of vector maps in image generation, MAGC (w/o SPADE block) can generate clear edges of ground objects, but the shapes of houses are still inaccurate.

\begin{table}[htb]
\setlength\tabcolsep{0.85pt}
\renewcommand{\arraystretch}{1.3}
\caption{Ablation study on the introduction of vector maps with various methods.}
\label{table3}
\resizebox{0.48\textwidth}{!}{
\begin{tabular}{ l  c  c c c}
\toprule
Methods & BD-PSNR$\uparrow$ & BD-LPIPS$\downarrow$ & BD-DISTS$\downarrow$ & BD-mIoU$\uparrow$\\
\hline 
MAGC(w/o map) & -3.0385 & -0.3452 & -0.1430 & 0.3349\\
MAGC(w/o SPADE block) & -3.0152 & -0.3429 & -0.1403 & 0.3353\\
MAGC & -2.9213 & -0.3521 & -0.1464 & 0.3567\\
\hline 
HiFiC  & -2.0506 & -0.3394 & -0.1152 & 0.2200 \\
HiFiC + map  & -2.1373 & -0.3443 &-0.1256 & 0.2503\\
\hline 
ELIC  & 0.3701 & -0.0429 & -0.0100 & 0.0646 \\
ELIC + map  & 0.3524 & -0.0479 &-0.0128 & 0.0788 \\
\bottomrule
\end{tabular}
}
\end{table}

\begin{figure*}[htb]
\centering
\includegraphics[width=\textwidth]{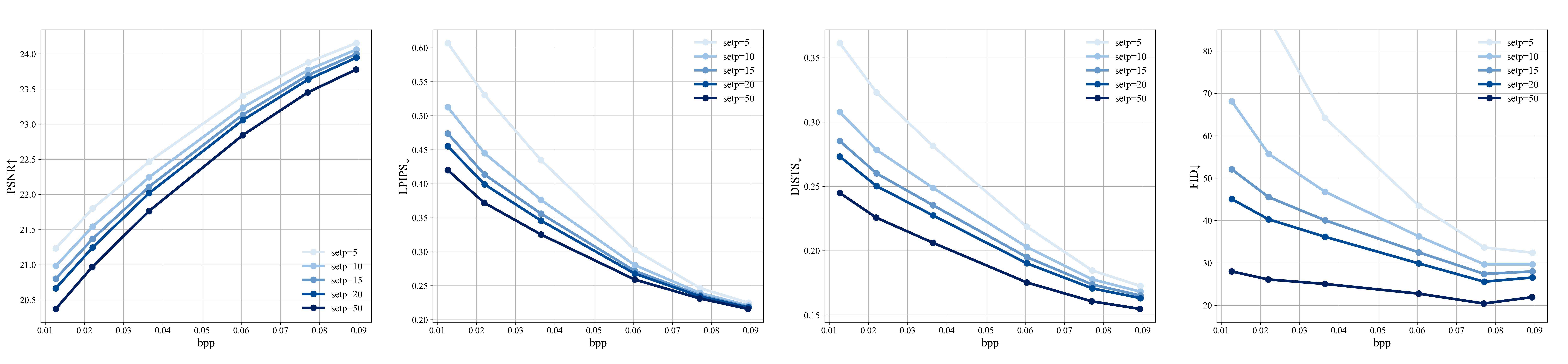}
\captionsetup{singlelinecheck=false} % 左对齐
\caption{Perception and distortion metrics depending on the number of denoising steps. In all cases these models are trained using a grid of 1,000 timesteps.}
\label{fig13} 
\end{figure*}

To explore the possibility of introducing vector maps in other methods, we choose the classical GAN-based HiFiC and VAE-based ELIC and inject semantic information from vector maps into them. Similar to the design of SAM, we utilize an adapter to produce multi-scale features which are then added to the feature maps at each scale on the decoding side. To compare the compression performance at the same bitrates, we fix the encoder and entropy model, only optimizing the decoder and adapter jointly. BD-metrics of these methods are shown in Table \ref{table3}. For both HiFiC and ELIC, integrating vector maps as semantic guidance results in lower PSNR scores, while perceptual metrics like LPIPS, DISTS and mIoU show varying degrees of improvement. Since perceptual quality is paid more attention at extremely low bitrates, we speculate that the introduction of vector maps should enhance the compression performance across various methods in RS image compression by designing more complex and reasonable modules.

\subsection{Impact of Denoising Steps}
PSNR, LPIPS, DISTS and FID results depending on the number of denoising steps are shown in Fig.\:\hyperref[fig13]{13}. We find that the metrics are comparatively constant across different denoising steps for higher bitrates, indicating that the diffusion model can reconstruct images with superior perceptual quality using a few denoising steps in these scenarios. Besides, we note that there is a trade-off between perception and distortion by changing the timesteps. Specifically, with the increase of denoising steps, our method can achieve better LPIPS, DISTS and FID scores, but pixel-level fidelity is worse in terms of PSNR.

% \subsection{Limitations}
% The limitations of the proposed MAGC can be summarized as follows: 1) Although the vector maps can be treated as additional semantic guidance, the map elements are limited in some main ground objects, such as roads, houses, rivers, and forests. For RS images with other objects, the introduction of vector maps has little effect. 2) Due to the limitations of the pre-trained autoencoder, our method exhibits poorer pixel-level fidelity in terms of PSNR and SSIM than baseline methods. 3) Our method requires more computational resources for training and longer time to reconstruct images compared to other VAE-based methods and GAN-based methods due to the iterations of the denoising diffusion process.

%%%%%%%%%%%%%%%%%%%%%%%%%%%%%%%%%%%%%%%%%%%%%%%%%%%%%%%%%%%%%%%%%%%%%%%%%%%%%%%%%%%%
%%%%  Conclusion
\section{Conclusion}\label{section6}
In this paper, we propose a novel SD-based image compression framework, employing a pre-trained diffusion model with powerful image priors and generative capability to reconstruct realistic RS images at extremely low bitrates. To produce semantically accurate reconstructions, we introduce vector maps as semantic guidance and present an improved image compression approach named Map-Assisted Generative Compression (MAGC). The image is first mapped into a latent representation via the pre-trained SD VAE encoder, and the latent representation is further compressed in LCM, serving as implicit guidance for the denoising diffusion process. The SAM is designed to extract multi-scale features from vector maps, providing explicit guidance for image generation. Quantitative and qualitative comparison results show that images produced by our method have the best perceptual quality compared to standard codecs and other learning-based methods. Besides, the proposed MAGC achieves the highest mIoU scores on the semantic segmentation task, demonstrating the superior practicality of our method.

In the future study, we will continue to explore generative compression methods on RS images. Based on the proposed method, we would like to find a strategy to achieve the balance between pixel-level fidelity and perceptual quality, and the denoising steps should also be considered to enhance efficiency in practical scenarios.

{
    \small
    \bibliographystyle{ieeenat_fullname}
    \bibliography{ref/bibfile}
}


\end{document}

%% file: preamble.tex
\usepackage{multirow} % 导入 multirow 宏包
\usepackage{makecell} % 导入 makecell 宏包